# Real-Time EEG Cap Electrode Detection for Guided Point-of-Care Placement

William Lehn-Schiøler[1–3,†], Mads Sverker Nilsson[3,†], Nicki Skafte Detlefsen[3]

[1] BrainCapture, Kongens Lyngby, Denmark
[2] Department of Health Technology, Technical University of Denmark, Lyngby, Denmark
[3] Department of Applied Mathematics and Computing, Technical University of Denmark, Lyngby, Denmark
[†] These authors contributed equally

Corresponding author: William Lehn-Schiøler (wls@braincapture.dk)

**Abstract**

We present a two-stage vision system that detects EEG cap electrodes in a live webcam stream and validates their anatomical placement in real time. A single-class YOLO detector localises electrodes; a geometric stage assigns each detection to a named 10–20 role from facial landmarks. Evaluating under subject-disjoint leave-one-subject-out (LOSO) cross-validation across five subjects wearing the clinically-validated Small/Medium/Large caps, the detector attains mAP@.5 $= 0.94 \pm 0.07$ across five held-out folds (0.96 pooled). A dedicated leave-one-cap-out axis, holding out every frame of a cap regardless of subject, leaves Medium and Large mAP@.5 within 0.01 of LOSO (0.97, 0.97) while Small drops to $0.72 \pm 0.28$, a gap confounded with subject familiarity rather than cap style. Geometric augmentation (rotation, perspective, mixup) improves in-plane-roll robustness and temporal-electrode recall at no inference cost, and a landmark-driven head crop extends the usable distance range, lifting mAP@.5 from 0.23 to 0.45 at $0.6\times$ apparent scale. A compact mobile-candidate backbone (YOLOv10n) keeps the detector at real-time throughput ($\sim 19$ FPS) on a commodity CPU at 640 px.

## 1. Introduction

EEG is the primary diagnostic and monitoring tool across a wide range of neurological conditions like epilepsy and non-convulsive status epilepticus, encephalopathies, coma and post-arrest prognostication, and sleep disorders among them; and closing the specialist gap that limits its use is a stated global health priority [1]. That gap is sharpest in low- and middle-income countries, where neurology and neurophysiology capacity is scarce [2], [3], [4], but it recurs wherever EEG is needed outside routine clinic hours: rapid-response and ICU EEG is increasingly used to rule out non-convulsive seizures in patients with altered consciousness, precisely when a dedicated EEG technologist is least likely to be on site [5], and longer ambulatory recordings [6] multiply the number of unsupervised donning events per workup. In each setting, whoever is available — an on-call resident, a ward nurse, a non-specialist operator — ends up placing the cap, and its diagnostic value (increasingly extended by automated, AI-assisted interpretation [7], [8], [9], [10]) depends on getting that placement right: electrodes away from their correct 10–20 positions [11] degrade the recorded signal directly and propagate into dipole/source-localisation error [12], [13], so no amount of downstream processing can recover a cap that was donned incorrectly. Low-cost, portable EEG devices are extending recording capability into exactly these settings [14], but doing so shifts placement onto whoever is on hand rather than a trained technologist, and even among experts, manual identification of standard 10–20 sites is unreliable across examiners [15], [16]; a non-specialist, unsupervised operator is a considerably harder case still. Placement quality assurance has not kept pace with placement accessibility, and closing that gap is the target of this work.

Existing approaches to verifying placement split into offline measurement and real-time guidance, and none combine camera-only hardware, real-time operation, and closed-loop verification of the electrodes actually placed. [16] validate electrode-position accuracy post-hoc with an optical scanner, offering no feedback during donning itself. Real-time guidance systems instead require additional hardware: a depth camera for 3D facial surface registration [17], or marker-based AR for a teaching aid [18]. Closest to our approach, [19] track monocular facial landmarks but predict **target** electrode positions from an anatomical atlas rather than observing the electrodes themselves, so a cap that has drifted after donning goes undetected. What is missing is a system that runs on the RGB camera already in any laptop or phone, in real time, and verifies where the electrodes physically are rather than where they are assumed to be.

Contributions:

1. A two-stage, real-time architecture: electrode detection followed by landmark-conditioned geometric role assignment.
2. A leakage-free, subject-disjoint (leave-one-subject-out) evaluation, paired with a dedicated leave-one-cap-out axis that separates cap-appearance generalisation from subject familiarity.
3. A controlled robustness study (lighting, distance, in-plane roll) via reproducible synthetic perturbations, isolating the adaptive head crop as the decisive factor at distance.
4. A throughput characterisation across input resolution and hardware.

## 2. Related Work

**Electrode localisation and placement.** Prior work on electrode and headgear placement splits into offline accuracy measurement and real-time placement guidance. On the measurement side, [16] compare an optical head scanner against photogrammetry for post-hoc electrode-position accuracy, and [15] show that even manual identification of a single standard site (C3/C4) is unreliable across examiners; both motivate why placement accuracy matters but offer no feedback during donning itself.

On the guidance side, [17] register a live RGB-D scan of the facial surface to a reference scan via iterative closest point (ICP) and overlay the resulting transform in augmented reality (AR); this targets repeatable repositioning relative to a subject's own earlier session rather than absolute 10–20 correctness, and needs a depth camera rather than a standard webcam. [18] use marker-based AR (fiducials tracked by Vuforia) to teach electrode placement, prioritising pedagogy over placement accuracy or hardware simplicity.

Closest to our approach, [19] track monocular facial landmarks and apply a pre-computed, atlas-derived transform, built from a library of over 1,000 head models, to estimate cranial 10–20 landmarks (nasion, preauricular points) and render them in AR, reaching a median position error of 0.75–1.52 cm depending on how subject-specific the atlas is. Like us, they use a single RGB camera and facial landmarks as the anatomical reference; unlike us, their system predicts where the electrodes should go from an anatomical atlas and never observes the electrodes themselves, so it cannot detect a cap that has since drifted from that target. Our detector instead localises the physical electrodes already on the head and checks them directly against the landmarks, closing a loop that atlas-only guidance leaves open. Across all four systems, none detects electrodes as a distinct visual class, and only [17] and [18] provide real-time visual feedback comparable in spirit to ours.

**Real-time object detection.** Single-stage detectors in the YOLO family [20] trade a small accuracy cost for inference speeds compatible with a live video feed, and pretraining on large natural-image corpora such as COCO [21] lets them fine-tune on small, task-specific datasets; a fit for our five-subject regime (Section 3). We use YOLOv8 [22] as the base detector; the same family has since produced markedly smaller variants, YOLOv10 [23] and YOLO11 [24], aimed at edge and mobile deployment, which we evaluate directly as candidates for the on-device pathway (Section 5.6). We treat detection as single-class: rather than an end-to-end model that jointly detects and names each of the five electrode roles (a five-class or keypoint formulation), an electrode-agnostic detector paired with a deterministic geometric stage (Section 4.2) avoids needing per-class training examples for every role at the cost of making role assignment dependent on facial-landmark geometry rather than learned appearance.

**Facial landmarks as anatomical anchors.** MediaPipe FaceLandmarker [25] provides real-time, on-device facial landmark tracking from a single RGB camera, without a depth sensor or fiducial markers: the same practical basis [19] use for atlas-based cranial-landmark estimation. We use it differently: rather than predicting target electrode positions from an anatomical atlas, its landmarks only establish a subject-specific scale (the inter-eye distance $d_{\text{ref}}$) and reference frame against which the directly detected electrodes are classified into roles and located for cropping (Section 4.2–4.3); the landmarks anchor the coordinate system rather than stand in for the electrodes themselves.

## 3. Dataset

The dataset is 1,309 webcam frames from 5 subjects, one continuous recording per subject-cap combination (subjects who wear more than one cap contribute one recording per cap), each frame carrying the full set of 5 electrode boxes (6,545 instances). Recordings were made with a laptop's built-in or USB webcam, one continuous take per subject wearing the BC-1 EEG cap [14], with frames sampled at a fixed interval from each video. Each frame was manually annotated by a single rater, who marked the pixel location of each of the 5 electrodes (F9, FP1, Fz, FP2, F10) with a point-and-click tool; each point was then converted to a fixed-size bounding box, which is why boxes are uniform squares rather than tightly fit to the visible electrode (Section 5). The study was conducted under the same DTU Compute Institutional Review Board approval (COMP-IRB-2026-01) as the companion clinical validation study, in accordance with the Declaration of Helsinki, with written informed consent obtained from all participants.

**Evaluation protocol.** We partition by subject, evaluating under leave-one-subject-out (LOSO): each fold trains on four subjects and validates on the held-out fifth. We additionally report a dedicated leave-one-cap-out axis (Section 5.3) that holds out every frame of a cap regardless of subject: a strictly harder test of cap-appearance generalisation, isolated from subject identity. Aggregate numbers are pooled across the five held-out folds, so no frame is scored by a model trained on it.

| Cap (size / colour) | Subjects | Frames | Electrode instances |
|---|---|---|---|
| Medium (Blue) | 4 | 875 | 4,375 |
| Large (Grey) | 3 | 261 | 1,305 |
| Small (Pink) | 2 | 173 | 865 |
| **Total** | **5** | **1,309** | **6,545** |

Table 1: **Dataset composition by cap.** All three caps are now worn by at least two subjects, so leave-one-subject-out is a genuine subject-generalisation test for every cap size; Figure 5 additionally reports a leave-one-cap-out axis.

## 4. Method

The system is a two-stage pipeline. Stage 1 is a class-agnostic electrode detector; Stage 2 assigns each detection to a named 10–20 electrode role using facial landmarks as anatomical anchors. Only Stage 1 is a learned model; Stage 2 is deterministic and shares its formulation with the deployed guidance system. See the pipeline diagram (Figure 1) for an overview.

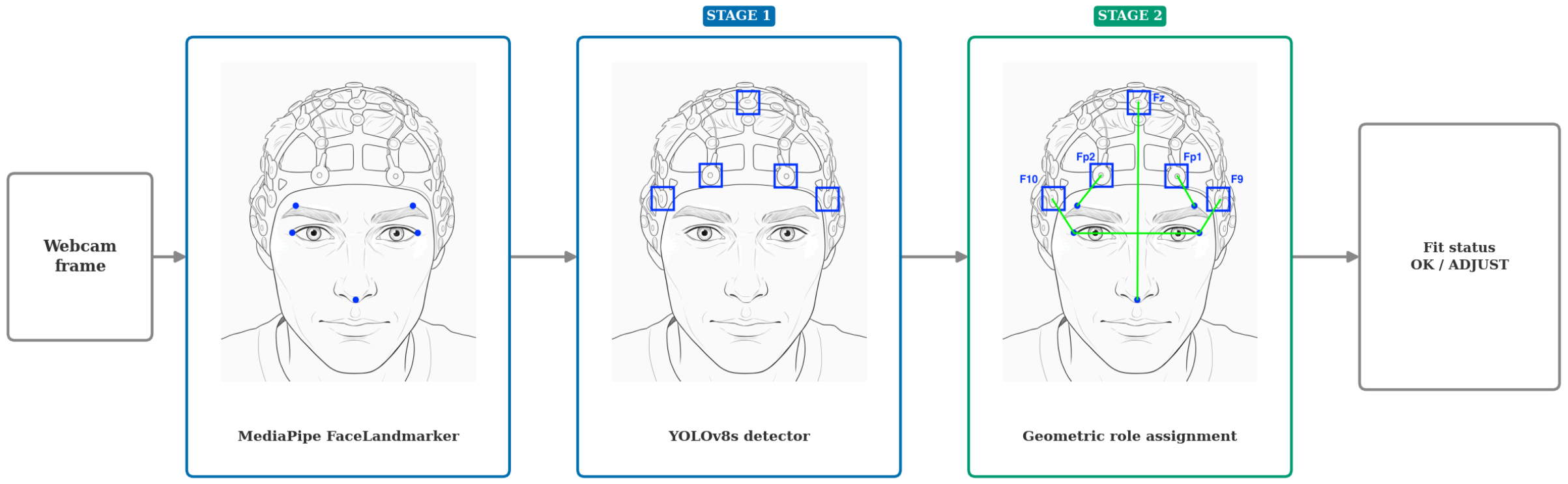

Figure 1: **Processing pipeline.** A webcam frame is passed to MediaPipe FaceLandmarker, whose five anchor points both drive the adaptive head crop and serve as the anatomical reference frame for role assignment. The YOLO detector (Stage 1) runs on the cropped frame; the top five detections are kept. Geometric role assignment (Stage 2) combines these detections with the landmarks to label each electrode and derive a per-electrode fit status.

### 4.1. Stage 1: Electrode detector

The detector is a YOLOv8s backbone trained at $960 \times 960$, using geometric augmentation (rotation (10°), perspective (0.0005) and mixup [26] (0.1)) on top of the deployed baseline recipe (HSV jitter 0.015/0.7/0.4, scale/translate 0.5/0.1, horizontal flip 0.5, mosaic 1.0)); an ablation (Section 5.7) confirms adding the three geometric terms improves accuracy and robustness at no inference cost. All results in this paper use this configuration.

### 4.2. Stage 2: Geometric role assignment

Detected electrodes are assigned to named roles (FP1, FP2, F9, F10, Fz) from facial landmarks by a deterministic geometric procedure, reproduced here in full since it is evaluated directly in Section 5.4. F9/F10 extend the standard 10–20 array [11] via the 10–10 combinatorial nomenclature [27]; the remaining three (FP1, FP2, Fz) are original 10–20 sites. All positions are normalised by the inter-eye reference distance $d_{\text{ref}} = \|\boldsymbol{e}_L - \boldsymbol{e}_R\|_2$, and a detection $\boldsymbol{d}$'s vertical offset from the eyebrow midpoint is $\delta_y(\boldsymbol{d}) = \left(d_y - y_{\text{brow}}\right)/d_{\text{ref}}$ (image $y$ increases downward, so negative $\delta_y$ is above the eyebrows). A detection is a temporal-electrode candidate (F10 on the image-left, F9 on the image-right) if it lies beyond a horizontal margin $\tau = 0.15 \cdot d_{\text{ref}}$ past the corresponding outer eye corner and $\delta_y > -0.5$ (excluding high-crown detections that would otherwise read as temporal); F10/F9 are then the left-/right-most candidates passing that test. Every remaining detection is an inner electrode, assigned by greedy nearest-neighbour proximity to the eyebrow landmarks: FP1 to the closest remaining detection to the image-right eyebrow, FP2 to the closest remaining detection to the image-left eyebrow, and Fz to whichever inner detection remains topmost.

### 4.3. Adaptive head crop

Before detection, the frame is cropped to the head using the facial landmarks from Stage 2. We take the bounding box of five stable landmarks (outer eyes, nose, inner eyebrows), expand it by $1.8\times$ horizontally and $2.5\times$ vertically to cover the cap, and clamp it to the frame. YOLO then internally upscales this smaller crop to its 960 px input, raising the effective pixel density over each electrode; detections are mapped back to full-frame coordinates by adding the crop origin. When no face is found the full frame is used. The crop is the single most important factor for detection at distance, where the cap otherwise occupies too few pixels (Section 5.2, Figure 4).

## 5. Experiments and Results

All experiments run the detector (recommended configuration, Section 4.1) through Ultralytics 8.4 (PyTorch 2.13) on an NVIDIA T4. Task-specific metrics (P, R, F1, all-electrodes-found rate) are computed at the operating confidence 0.25 by greedy one-to-one IoU matching at threshold 0.5; mAP@.5 and mAP@.5:.95 come from the Ultralytics validator. Because the boxes are fixed-size squares by annotation convention, mAP@.5 is the meaningful detection metric; the high-IoU

mAP@.5:.95 largely reflects that fixed box size. mAP is also the more reproducible figure: it integrates over every confidence threshold, whereas P/R/F1 are read at one fixed cutoff (0.25) and can shift by a few points between runs as ordinary floating-point non-determinism in GPU inference moves borderline-confidence detections across that single line. Treat mAP@.5 as accurate to the reported digits and P/R/F1 as indicative of the operating point rather than exact.

## 5.1. Detection performance

Every cap size is now worn by at least two subjects, so all five LOSO folds are a genuine subject-generalisation test: they average $0.94 \pm 0.07$ mAP@.5 (Table 2), with fold s4 (subject 4, Small cap) the weakest at 0.80, a per-subject outlier (Section 5.5) rather than a cap effect, since subject 5 also wears Small and scores in line with the rest (0.95, fold s5). Pooling every prediction (frame-weighted) gives 0.964 mAP@.5 (Figure 2); the F1–confidence curve peaks at 0.94 near confidence 0.41, so the deployed 0.25 threshold is slightly recall-favouring.

| **Held-out fold** | **Frames** | **P** | **R** | **mAP@.5** | **All-5** |
|---|---|---|---|---|---|
| Subject 1 | 610 | 0.859 | 0.990 | 0.990 | 0.961 |
| Subject 2 | 190 | 0.942 | 0.952 | 0.980 | 0.779 |
| Subject 3 | 136 | 0.824 | 0.985 | 0.991 | 0.926 |
| Subject 4 | 71 | 0.861 | 0.713 | 0.797 | 0.070 |
| Subject 5 | 302 | 0.959 | 0.930 | 0.954 | 0.738 |
| **Mean ± std** | 1,309 | 0.889±0.052 | 0.914±0.103 | **0.943±0.074** | 0.695±0.324 |
| **Pooled** | 1,309 | – | – | **0.964** | 0.831 |

Table 2: **LOSO detection performance.** Per-fold results across all five held-out folds; "All-5" is the fraction of frames in which every electrode is matched. The mean row is equal-weight across folds; the pooled row is frame-weighted.

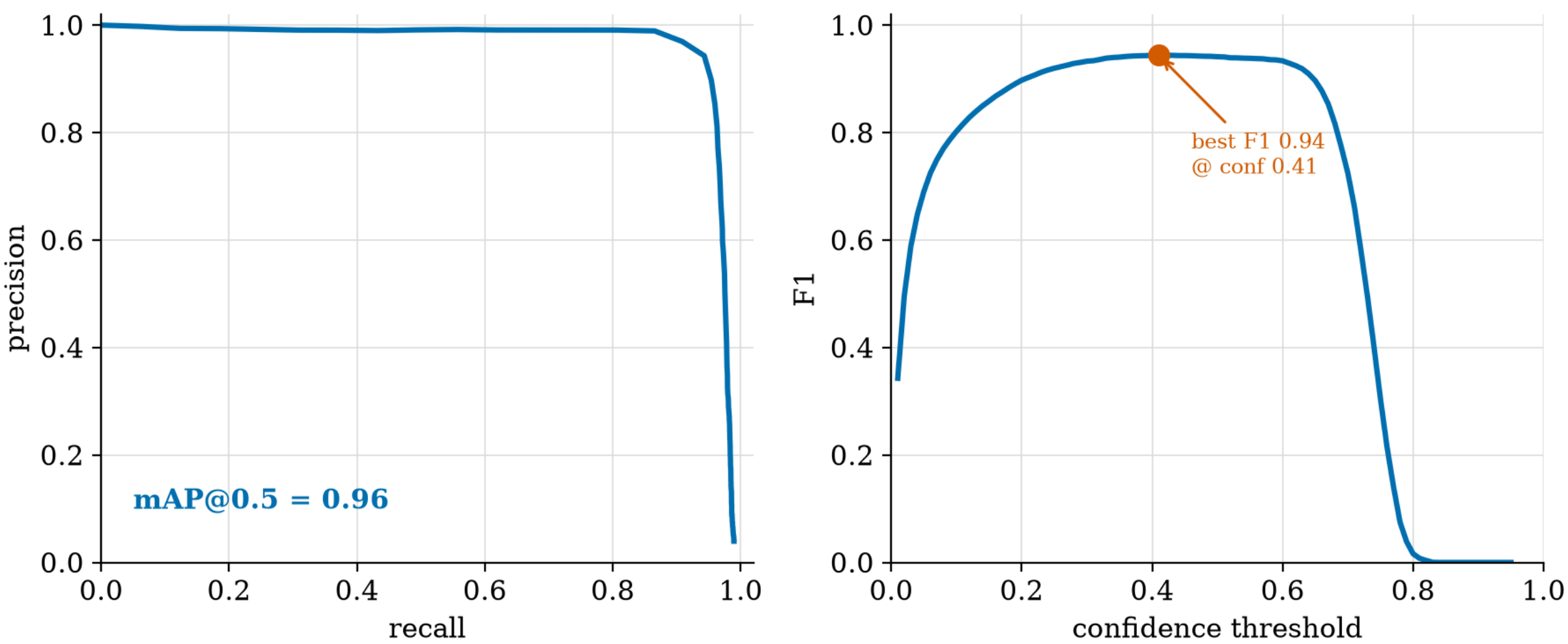

Figure 2: **Detector operating characteristics.**
Left: precision–recall (area = mAP@.5 = 0.964). Right: F1 vs.confidence, peaking at 0.94 near confidence 0.41.

Beyond standard detection metrics, each frame is expected to contain a fixed set of 5 electrodes. The subject-generalisation all-electrodes-found rate is 0.70 with a mean detection count close to the ground-truth 5 per frame; the residual gap is concentrated at specific electrode positions rather than spread uniformly (Section 5.5), and heavily at one fold: Subject 4′s vertex electrode (Fz) is missed in 93% of that fold's 71 frames (vs. 1–21% elsewhere), which alone accounts for its 0.07 all-5 rate (Table 2). Excluding that single small-sample fold, the remaining four folds average $0.85 \pm 0.09$; we report the full five-fold number as the headline result and treat the exclusion only as a sensitivity check (Section 6).

### 5.2. Robustness ablations (synthetic)

All robustness experiments apply reproducible, single-variable perturbations to the held-out frames of all five LOSO folds. They are synthetic proxies (gamma and scale transforms, not captures under real varied conditions) and are reported as such; the caveat is restated in the limitations. As noted above, the F1 values below are read at a single fixed confidence threshold and are indicative rather than exact to the last digit.

**Lighting.** We sweep three standard, independent tone transforms well past the mild range used at train time, out to where detection actually breaks (Figure 3): gamma (a nonlinear brightness curve, $I_{\text{out}} = I_{\text{in}}^{1/\gamma}$; $\gamma < 1$ brightens, $\gamma > 1$ darkens, $\gamma = 1$ is unperturbed), brightness (a linear gain multiplying every pixel), and contrast (a linear scaling of pixel values about mid-grey). F1 stays within $\sim 0.02$ of the unperturbed baseline ($0.90 \pm 0.06$) for $\gamma \in [0.6, 2.2]$ and brightness/contrast $\times [0.5, 1.5]$ (consistent with the training recipe's HSV jitter, Section 4.1), but degrades sharply beyond that: severe darkening ($\gamma = 0.15$) collapses to F1 $0.38 \pm 0.23$, over-brightening ($\times 2.5$) to $0.53 \pm 0.16$, and heavy contrast ($\times 3.0$) to $0.79 \pm 0.07$.

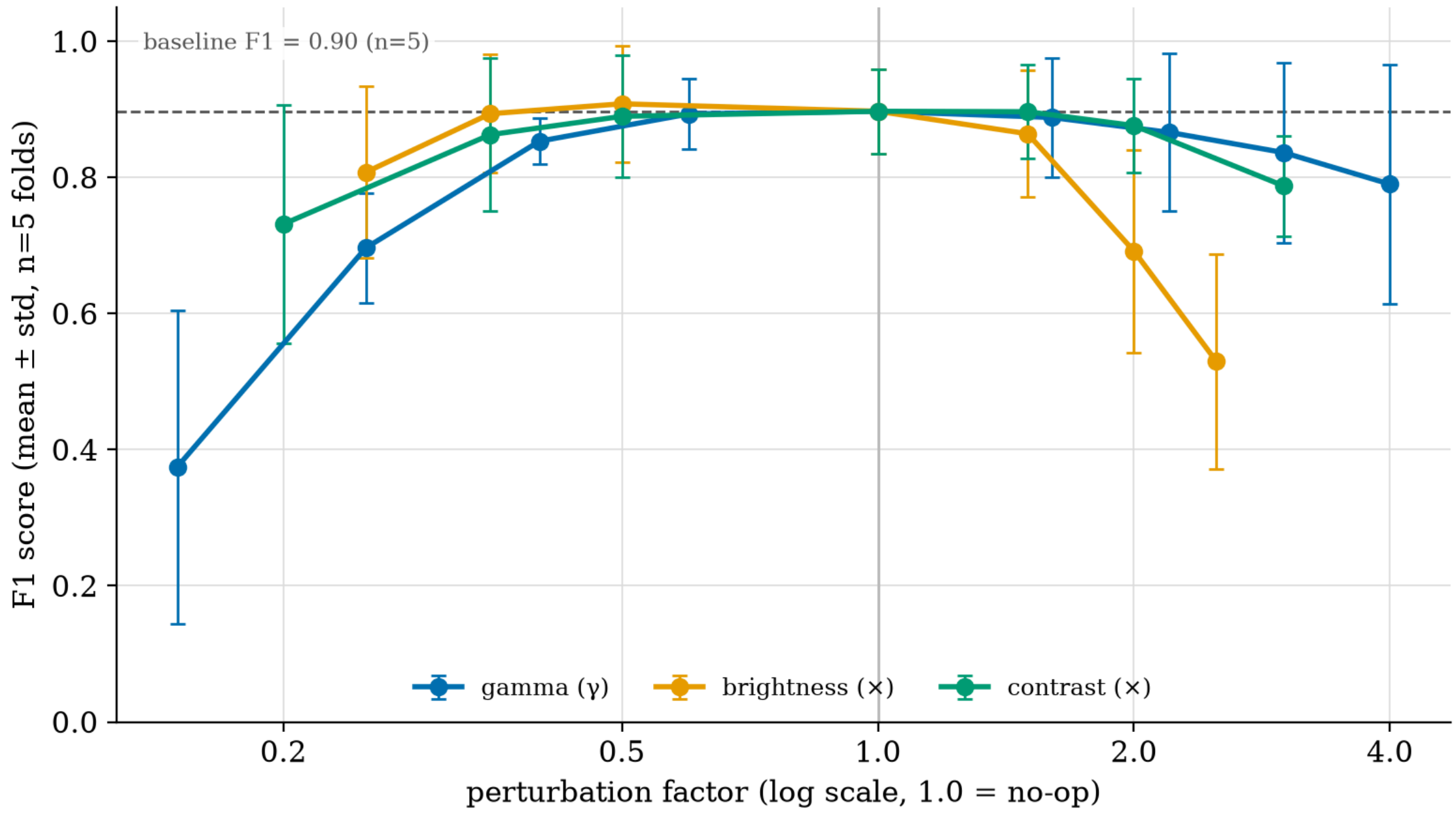

Figure 3: **Lighting robustness.** F1 (mean $\pm$ std across all five folds) vs. perturbation strength (log scale) for gamma, brightness and contrast. All three hold near baseline across a wide middle range and fall off steeply at the extremes.

**Distance.** Subject scale is reduced synthetically (shrink-and-pad) so the pixel inter-eye distance $d_{\text{ref}}$ decreases, emulating a more distant subject. The headline experiment contrasts detection **with vs. without** the adaptive head crop (Figure 4, left). Without the crop, mAP@.5 falls from $0.94 \pm 0.07$ to $0.23 \pm 0.24$ at $0.6\times$ scale and to $\sim 0.00$ by $0.4\times$; with the crop it holds better in that same middle range ($0.45 \pm 0.20$ at $0.6\times$, $0.01 \pm 0.02$ at $0.4\times$) but is not uniformly better: at $1.0\times$ scale the crop is actually somewhat below the full-frame baseline (0.80 vs. 0.94), most likely because cropping and upsampling an already-close subject changes the input even when no crop was needed. The crop's benefit is concentrated in the middle distance range, not universal. Neither survives $0.3\times$, where MediaPipe can no longer locate the face to crop.

**Roll.** An in-plane rotation. With rotation augmentation in the training recipe (Section 4.1), the detector holds F1 $0.86 \pm 0.10$ at $\pm 10°$ and $0.69 \pm 0.15$ at $\pm 20°$ (pooled over both directions and all five folds; the two directions are not symmetric); a recipe without rotation augmentation was far weaker here (Figure 4, right; ablation details in Section 5.7).

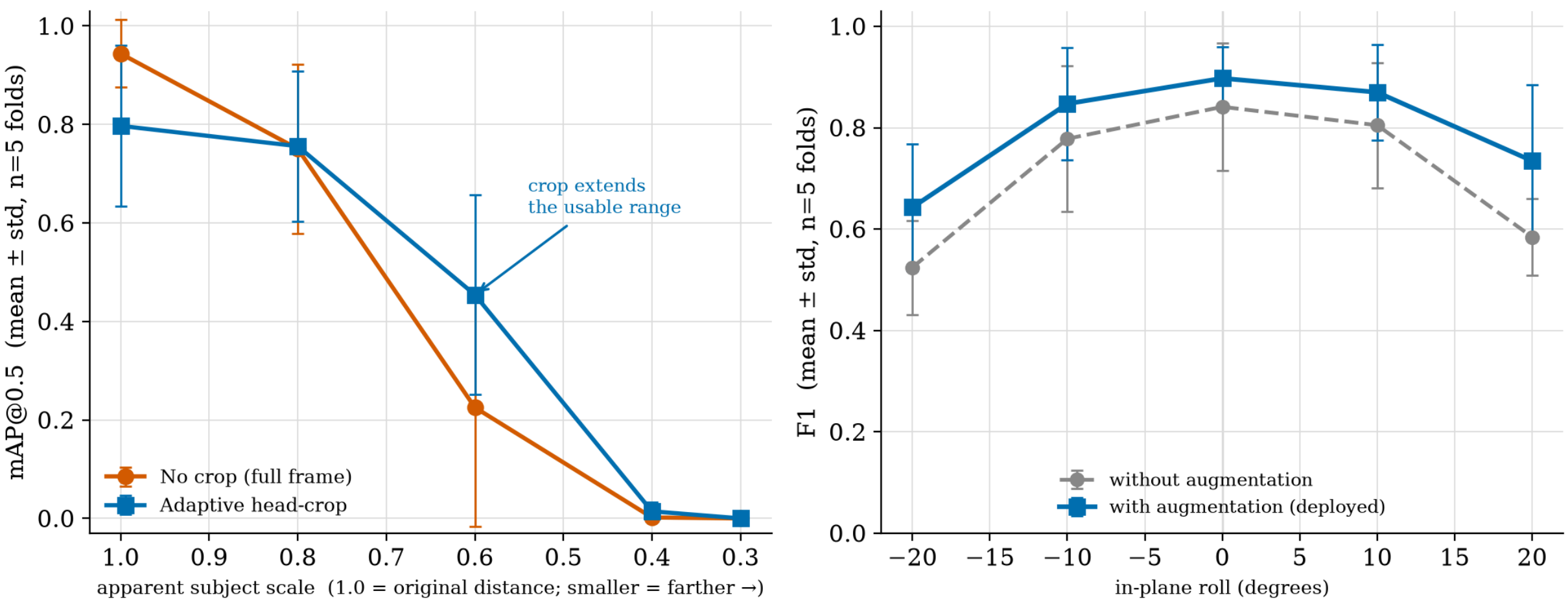

Figure 4: **Distance and roll robustness.** Mean ± std across all five folds. Left: mAP@.5 with vs. without the adaptive head crop; the crop's benefit is concentrated in the middle distance range (∼ 0.6 × scale) and is not uniform (see text). Right: F1 for in-plane roll, with vs. without the augmentation recipe (rotation, perspective, mixup), same five-fold scope, pooled over both roll directions.

| **Perturbation (F1, mean ± std)** | **none** | **mild** | **moderate** | **severe** |
|---|---|---|---|---|
| Lighting (gamma) | 0.90±0.06 | 0.89±0.09 | 0.89±0.05 | 0.85±0.03 |
| Distance, w/ crop | 0.82±0.10 | 0.80±0.12 | 0.58±0.14 | 0.07±0.05 |
| Distance, no crop | 0.90±0.06 | 0.80±0.13 | 0.33±0.22 | 0.02±0.03 |
| In-plane roll | 0.90±0.06 | 0.86±0.10 | 0.69±0.15 | – |

Table 3: **Robustness to synthetic perturbations.** Reported as F1 at confidence 0.25 (mean ± std across all five folds; roll additionally pools both ± directions, $n = 10$). Severity bins: lighting $\gamma = 1.0$ / 1.6 / 0.6 / 0.4; distance scale = 1.0 / 0.8 / 0.6 / 0.4; roll = 0 / ±10 / ±20°. Per-level mAP for the distance rows is in Figure 4; the wider lighting sweep is in Figure 3.

### 5.3. Cap size and colour

We stratify mAP@.5 by cap under two generalisation axes (Figure 5): **leave-one-subject-out**, where the cap itself may still be seen via other subjects wearing it, and **leave-one-cap-out**, where the model never sees any frame of that cap regardless of who wore it. With a second subject now wearing Small, all three caps have at least two subjects on both axes.

Leave-subject-out mAP@.5 is high and tight for all three: Medium 0.98 ± 0.01, Large 0.97 ± 0.01, Small 0.87 ± 0.06. Leave-cap-out (a strictly harder test, since the model has never seen that cap's size/colour at all) leaves Medium and Large nearly unchanged (0.97 ± 0.03, 0.97 ± 0.02), but Small drops sharply to 0.72 ± 0.28. That spread splits almost perfectly by subject (subject 5: 0.99, subject 4: 0.44): subject 5 wears all three caps, so the Small-out fold still trains on their Medium/ Large frames, while subject 4 wears only Small, making that fold simultaneously a new cap and a new identity. The gap is therefore confounded with subject familiarity rather than a clean read on cap-style generalisation alone (Section 6, ii).

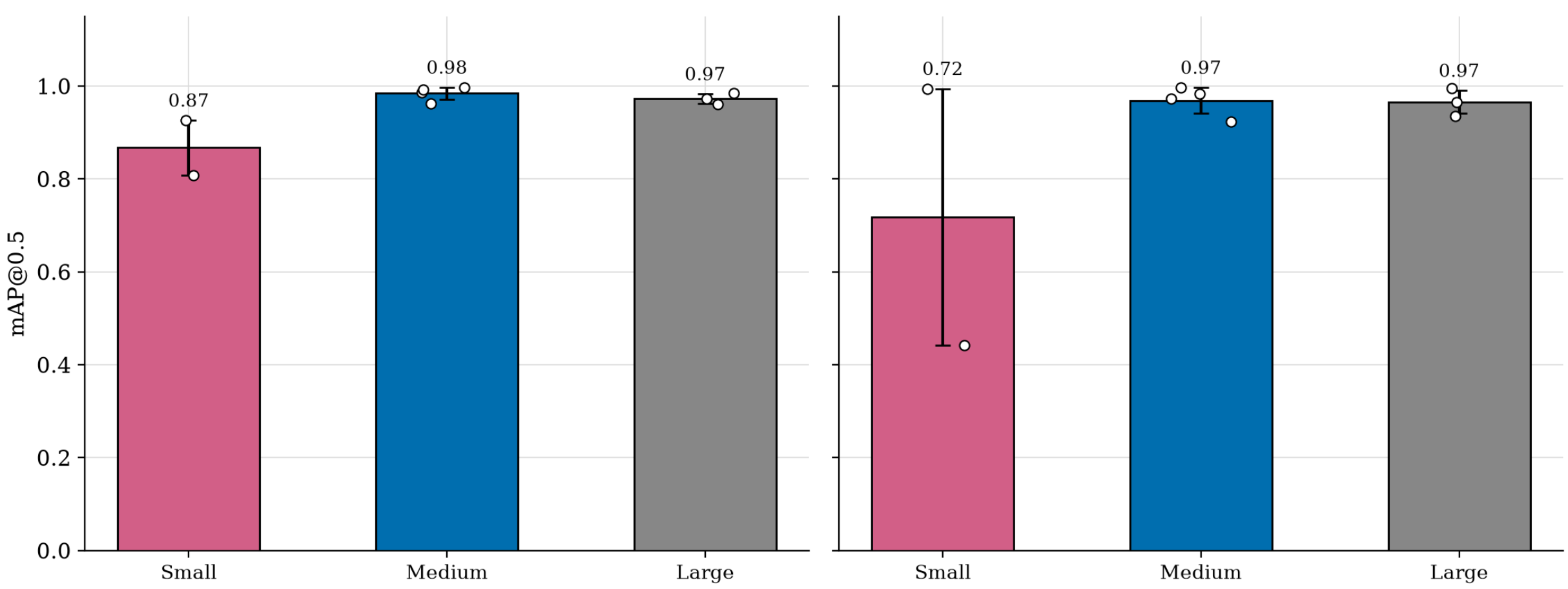

Figure 5: **Detection by cap size.** mAP@.5 under leave-one-subject-out (left) vs. leave-one-cap-out (right); colour is the cap's own colour, white dots are the individual subjects contributing to that bar (Medium $n = 4$, Large $n = 3$, Small $n = 2$). Error bars are mean ± std across those subjects.

### 5.4. Downstream role assignment

The product output is the named 10–20 role of each detection, assigned by Stage 2′s deterministic rule (Section 4.2). We evaluate it against the detector's actual predictions (sometimes missing a box, Fz especially, Section 5.5; sometimes noisy near threshold), reading each prediction's true label off its IoU-matched ground-truth box in x-sorted order, which is valid because the cap's layout is physically fixed and the dataset matches the system's assumed roughly-frontal pose (Section 6).

Pooled over all five folds, Stage 2 recovers the correct role for 90.7% of matched detections (4,361 / 4,807; Figure 6, right). FP1 (98.9%) and F10 (95.1%) are most reliable; Fz is weakest (79.9%), leaking mostly into FP2 (19.9%), consistent with its high miss rate (Section 5.5) starving the FP1/FP2/Fz proximity fallback. F10 and F9 leak smaller fractions into their neighbouring frontopolar role (4.9%, 10.2%); a further 2.0% (100 / 4,807) go unassigned rather than mislabelled (no usable landmarks or too few detections) and are excluded from the accuracy figure.

### 5.5. Failure analysis

Failures are not spread uniformly across electrodes. Figure 6 (left) reports the per-electrode miss rate for the final model as mean ± std across the five current LOSO folds, rather than a single pooled number, so the fold-to-fold spread is visible instead of hidden. The vertex electrode **Fz** dominates (13% pooled; 25% ± 35% by fold) and the temporal electrode **F9** is next widest (5% pooled; 11% ± 15% by fold); in both cases nearly all of that spread traces back to the fold s4 outlier (Section 5.1) rather than five comparably hard folds. F10 and the frontopolar pair FP1/FP2 stay low and tight across folds (∼3–4% and ∼1–2% respectively). The temporals sit farthest from the face centre and were the first to leave the frame under foreshortening; adding rotation and perspective augmentation (Section 5.7) largely fixed them. Fz remains the hard case: it is highest on the head and most affected by hair and by the top image border, and, as Section 5.7 shows, is not helped by higher input resolution; it is an occlusion/framing problem, not a pixel-density one.

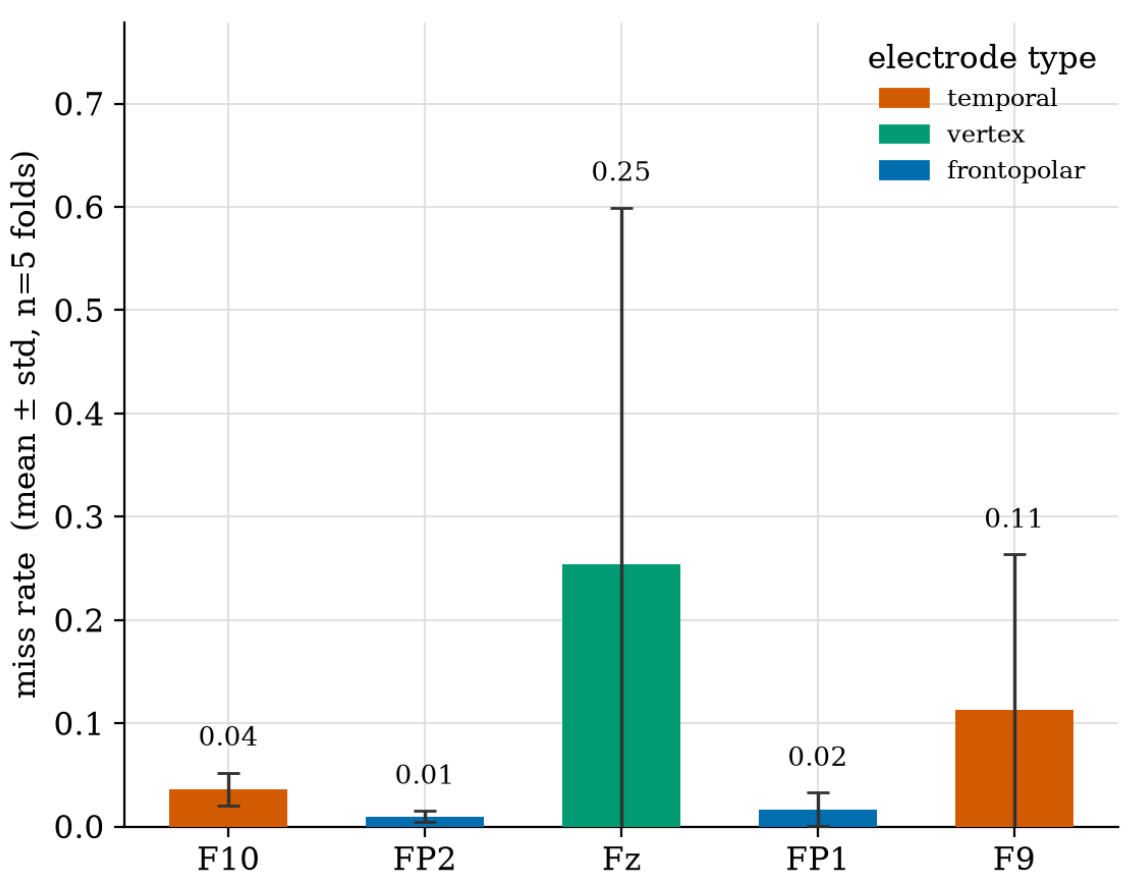

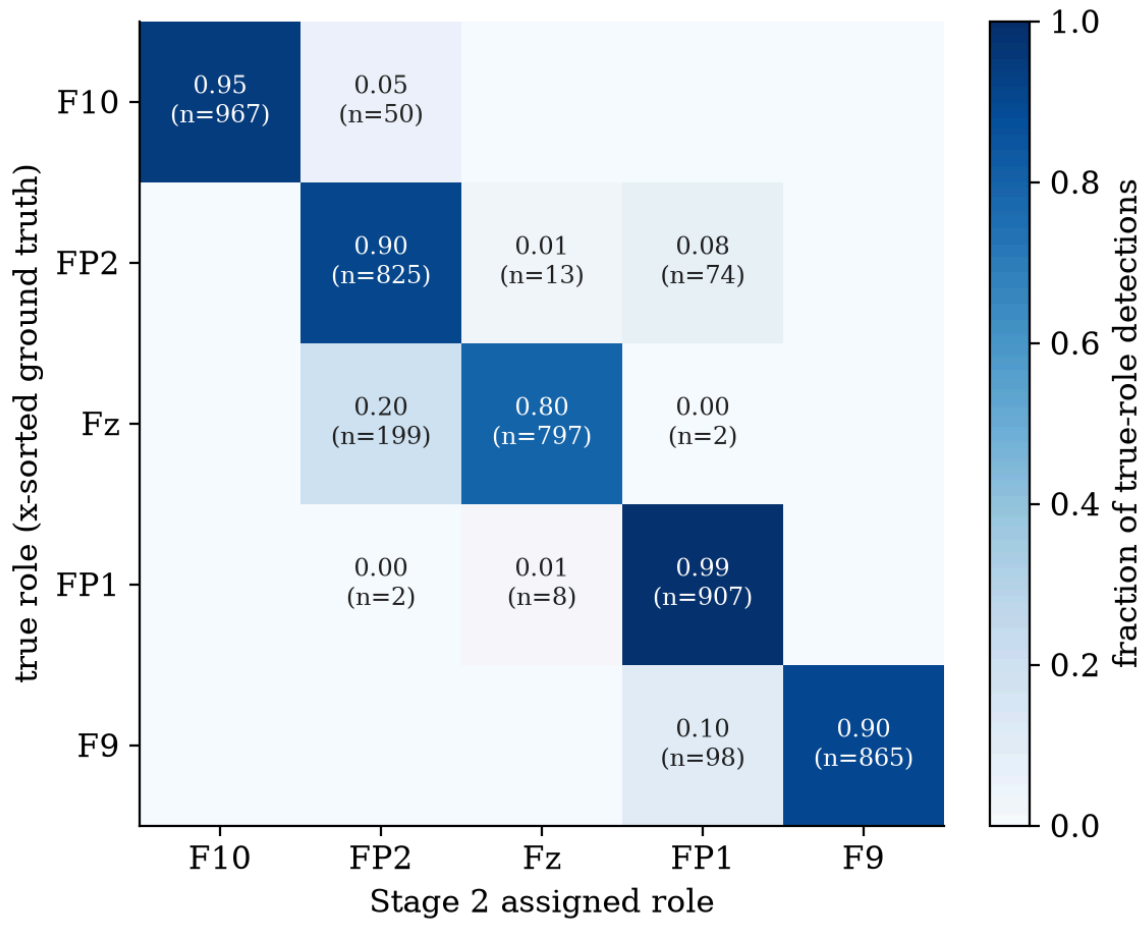

Figure 6: **Per-electrode miss rate and role-assignment confusion.** Left: miss rate for the final deployed model, mean ± std across the five current LOSO folds. Fz and F9′s wide bars trace to fold s4′s outlier performance (Section 5.1), not measurement noise. Right: role-assignment confusion matrix, row-normalised, pooled over all five folds; Fz's high miss rate (left) is what starves the FP1/FP2/Fz proximity fallback, leaking mostly into FP2 (Section 5.4).

### 5.6. Deployment and operational characteristics

The system runs inference server-side: the browser streams JPEG frames over a WebSocket to a Cloud Run server and gets JSON coordinates back, so the client (including mobile) stays thin, and user-visible latency is dominated by the network round-trip rather than the $\sim 24$ ms detector call. A single T4 sustains the full MediaPipe + crop + YOLO pipeline at $\sim 19$ FPS at 960 px (within the 15–30 FPS target) while serving multiple clients; the detector alone runs at 40 FPS, or 81 FPS at 640 px (Figure 7).

**Toward on-device deployment.** For low-connectivity point-of-care settings the system should eventually run on the phone itself, via two levers (Table 4). First, a smaller detector: sweeping every COCO-pretrained Ultralytics backbone below the default: YOLOv8n (3.0 M), YOLO11n (2.6 M), YOLOv10n (2.3 M), a fifth of the default's parameters, gives 0.72–0.86 mAP@.5 with wide, overlapping error bars (±0.19–0.29), so accuracy alone shouldn't decide this. Nearly all of that spread traces to fold s4 (subject 4, Section 5.1) — each architecture's single worst fold (Table 4) — where the models separate clearly without tracking size or recency.

We pick YOLOv10n for constrained deployment: the smallest footprint (2.3 M params, 5.8 MB) at speed comparable to YOLOv8n (99 vs. 108 GPU FPS), despite having fewer FLOPs (6.5 vs. 8.1 GFLOPs); its efficiency is architectural (depthwise-separable convolutions, a partial self-attention block, an NMS-free head) rather than a raw-compute win, so GFLOPs and FPS don't track each other here. YOLOv8s stays the default since its extra capacity should tell with more training data; a smaller-still detector would mean training from scratch without COCO pretraining, or a dedicated tiny-detector framework such as NanoDet [28] or PicoDet [29], left as future work. A mobile-native runtime: exporting to TFLite / Core ML / NCNN with a GPU/NPU delegate and static INT8 quantisation brings modern phones into the real-time band.

| Detector | Params | GFLOPs | Subj-gen mAP | Fold s4 | GPU FPS | CPU FPS |
|---|---|---|---|---|---|---|
| YOLOv8s | 11.1 M | 28.4 | 0.86 ± 0.19 | 0.48 | 81 | 11.4 |
| YOLOv8n | 3.0 M | 8.1 | 0.83 ± 0.24 | 0.35 | 108 | 19.0 |
| YOLO11n | 2.6 M | 6.3 | 0.72 ± 0.28 | 0.21 | 78 | 16.0 |
| YOLOv10n | 2.3 M | 6.5 | 0.81 ± 0.29 | 0.23 | 99 | 18.5 |

Table 4: **Detector variants at the mobile operating point.** 640 px, T4 GPU and x86 CPU; GFLOPs at 640 px; model sizes 22.6 / 6.3 / 5.5 / 5.8 MB in row order. "Subj-gen mAP" is mAP@.5, mean ± std across all five LOSO folds, the same subject-generalisation estimate as Table 2; "Fold s4" is each architecture's single lowest-scoring fold and is broken out separately since it is the largest driver of the spread. The deployed default runs YOLOv8s at 960 px instead (Table 2, Figure 7); **YOLOv10n** (last row) is the recommended mobile candidate.

Figure 7 plots the resulting speed–accuracy frontier: accuracy is bounded by resolution (mAP@.5 0.96 at 960 px, 0.92 at 640, collapsing below 480 px as electrodes fall below a few pixels), while speed is set by resolution, model and hardware. The knee sits around 640 px (∼ 96% of 960-px accuracy at roughly double the throughput). Only the GPU reaches the real-time band with the large detector; a smaller backbone brings even a CPU into that band at 640 px (0.91 mAP@.5): the recommended on-device operating point.

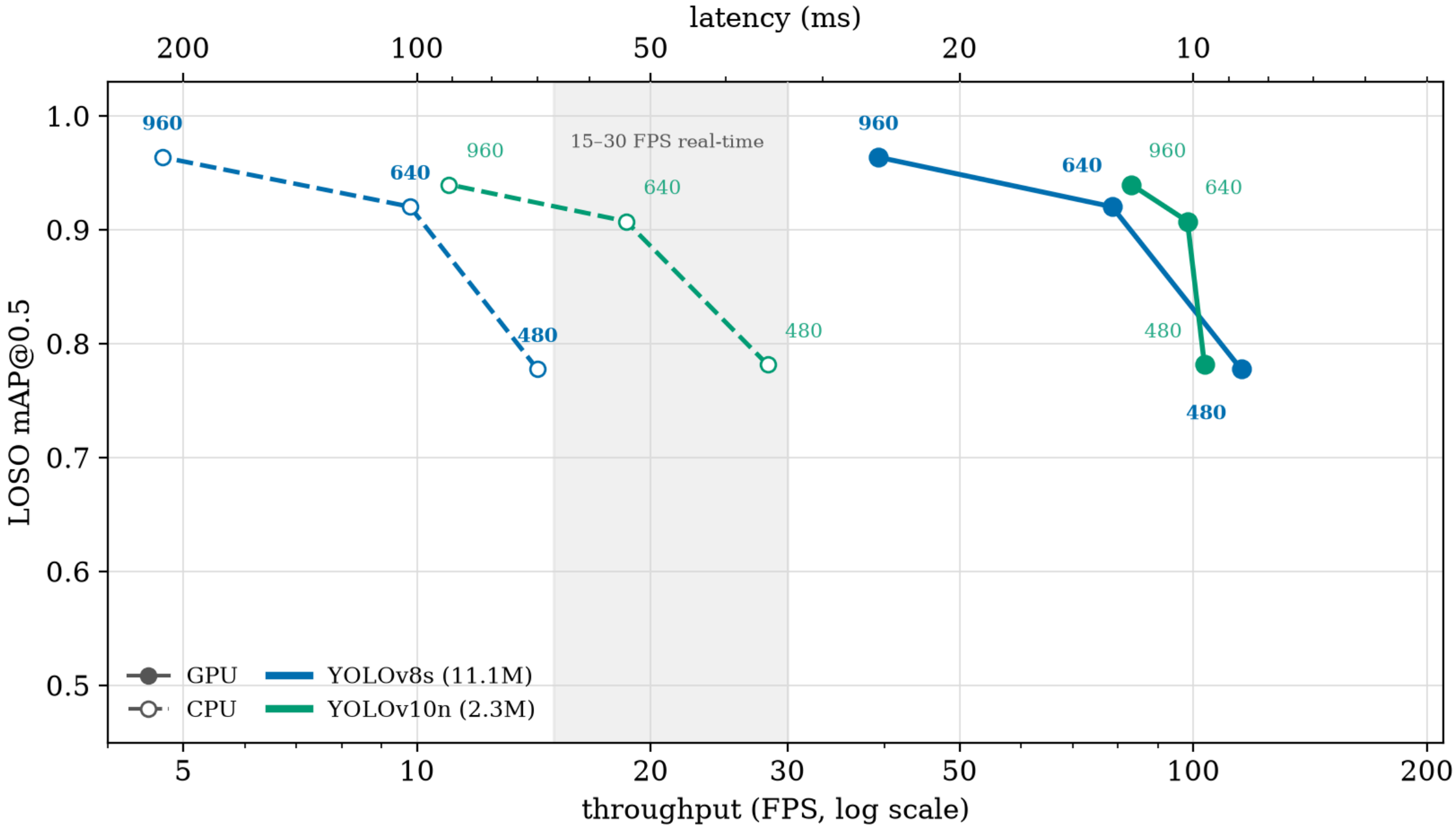

Figure 7: **Speed–accuracy frontier.** Over model (blue = YOLOv8s, 11.1 M params; green = YOLOv10n, 2.3 M), input resolution (point labels in px, an NVIDIA T4 GPU and x86 server CPU), and hardware (solid / filled = GPU, dashed / hollow = CPU). YOLOv8n is omitted: at every resolution it sits within a point of YOLOv10n (Table 4), so plotting both would only overlap two lines with no added information. Accuracy is essentially resolution-bound; YOLOv10n sits further right at close to the same mAP (faster at near-equal accuracy). Notably YOLOv10n at 640 px on CPU enters the 15–30 FPS real-time band (shaded), the on-device operating point. Resolutions below 480 px collapse (mAP < 0.2) and are omitted.

### 5.7. Detector augmentation and input resolution

The recommended configuration adds geometric augmentation (rotation 10°, perspective, mixup) to the deployed recipe, which trained without it (rotation = 0). We validate it with a same-scope roll ablation: a no-augmentation baseline retrained on the current five folds, matched against the deployed augmented checkpoints already used throughout this paper (Figure 4, right). Augmentation raises pooled roll-robustness F1 at no inference cost: at ±20° from 0.55 ± 0.09 to 0.69 ± 0.15, and at ±10° from 0.79 ± 0.13 to 0.86 ± 0.10 (mean ± std, pooled over both roll directions, same five folds as Table 2). Per fold, this holds in 4 of 5 folds at each angle (one fold reverses slightly at each angle, by 0.02 or less); with only five folds, a paired significance test has essentially no power to confirm this formally (the best-case two-sided $p$ at $n = 5$ is 0.0625), so we report the direction qualitatively rather than claim statistical significance.

## 6. Limitations

Several limitations bound these results.

**(i) Scale and diversity:** the study covers only 5 subjects (S/M/L caps), so LOSO folds are small (each trains on four subjects) and we report mean±std; subject, skin-tone, hair-type and cap-model diversity are all limited, and a single hard fold can dominate an aggregate metric rather than average out (Subject 4′s all-electrodes-found rate, Section 5.1). The intended fix is broader enrollment, not excluding that fold.

**(ii) Cap confounding:** A dedicated leave-one-cap-out axis (Section 5.3) shows Small splitting sharply by subject (0.99 vs. 0.44), which we attribute to subject familiarity rather than cap appearance; with only two subjects per cap on either axis, this is not yet a settled read.

**(iii) Synthetic perturbations:** the lighting, distance and roll sweeps are deterministic image transforms, not captures under real varied conditions, and may under- or over-state real robustness. Roll is swept only to ±20°, not to the same severe/collapse level as lighting and distance, because the synthetic ground truth itself becomes unreliable beyond that: perturbed boxes are relocated to the rotated centre but kept axis-aligned and the same size (the annotation convention), so at larger angles the box increasingly fails to bound the actually-rotated electrode, and F1 would conflate genuine detector failure with this annotation artefact rather than measure robustness cleanly.

**(iv) The Fz electrode** remains a hard case (13% pooled miss) that augmentation and higher resolution do not fix (Section 5.7).

**(v) Role accuracy:** Stage 2 recovers the correct role for 90.7% of matched detections against the detector's actual output (Section 5.4); the residual error concentrates in Fz, whose own detection is also the least reliable (Section 5.5).

## 7. Conclusion

We presented a real-time, camera-only system that detects EEG cap electrodes and validates their placement, evaluated under a leakage-free leave-one-subject-out protocol scoped to the clinically-validated EEG caps. Every cap size is worn by at least two subjects, so LOSO is a genuine subject-generalisation test across all three: the detector reaches $0.94 \pm 0.07$ mAP@.5 across the five held-out folds (0.96 pooled), with a dedicated leave-one-cap-out. Geometric augmentation in the deployed recipe removes an in-plane-roll blind spot and cuts temporal-electrode misses at no inference cost, and the landmark-driven head crop extends the usable distance range in the middle distance band (mAP@.5 $0.23 \to 0.45$ at $0.6 \times$ scale) while the full pipeline holds $\sim 20$ FPS on a commodity server GPU. Future work: broader subject and cap diversity (more subjects per cap, and more cap sizes).

---

**Data and code availability.** The code used for training and evaluation is not publicly released. Please contact the corresponding author for access to the code and dataset.

**Ethics.** Data collection was conducted under DTU Compute Institutional Review Board approval COMP-IRB-2026-01, in accordance with the Declaration of Helsinki, with written informed consent obtained from all participants.

**Competing interests.** W.L.S. is affiliated with BrainCapture, the distributor of the BC-1 cap evaluated in this work; W.L.S., M.S., and N.S.D. are affiliated with DTU. The authors declare no other competing interests.